\documentclass[11pt]{article}

\usepackage[utf8]{inputenc}
\usepackage[T1]{fontenc}
\usepackage{lmodern} 

\usepackage[a4paper,margin=1in]{geometry}
\usepackage{microtype}
\usepackage{setspace}

\usepackage{amsmath,amssymb,amsfonts,amsthm}
\numberwithin{equation}{section} 

\theoremstyle{definition}

\theoremstyle{remark}

\usepackage{graphicx}
\graphicspath{{figs/}} 
\usepackage{booktabs}
\usepackage{multirow}
\usepackage{caption}
\usepackage{subcaption} 
\usepackage{float} 

\usepackage[ruled,vlined]{algorithm2e} 

\usepackage{hyperref}
\hypersetup{
  colorlinks=true,
  linkcolor=blue,
  citecolor=blue,
  urlcolor=blue,
  pdfauthor={First Author},
  pdftitle={Paper Title}
}
\usepackage[nameinlink,capitalise,noabbrev]{cleveref}

\usepackage[numbers,sort&compress]{natbib}
\usepackage{listings}
\usepackage{authblk}
\title{Scaling Law for Large-Scale Pre-Training Using Chaotic Time Series and Predictability in Financial Time Series}
\author[1]{Yuki Takemoto\thanks{Corresponding author: u@hay.co.jp or X: @blog\_uki}}
\affil[1]{HAY Co., Ltd.}
\date{\today} 

\begin{document}
\maketitle

\begin{abstract}
Time series forecasting plays a critical role in decision-making processes across diverse fields including meteorology, traffic, electricity, economics, finance, and so on. Especially, predicting returns on financial instruments is a challenging problem. Some researchers have proposed time series foundation models applicable to various forecasting tasks. Simultaneously, based on the recognition that real-world time series exhibit chaotic properties, methods have been developed to artificially generate synthetic chaotic time series, construct diverse datasets and train models. In this study, we propose a methodology for modeling financial time series by generating artificial chaotic time series and applying resampling techniques to simulate financial time series data, which we then use as training samples. Increasing the resampling interval to extend predictive horizons, we conducted large-scale pre-training using 10 billion training samples for each case. We subsequently created test datasets for multiple timeframes using actual Bitcoin trade data and performed zero-shot prediction without re-training the pre-trained model. The results of evaluating the profitability of a simple trading strategy based on these predictions demonstrated significant performance improvements over autocorrelation models. During the large-scale pre-training process, we observed a scaling law-like phenomenon that we can achieve predictive performance at a certain level with extended predictive horizons for chaotic time series by increasing the number of training samples exponentially. If this scaling law proves robust and holds true across various chaotic models, it suggests the potential to predict near-future events by investing substantial computational resources. Future research should focus on further large-scale training and verifying the applicability of this scaling law to diverse chaotic models.
\end{abstract}



\vfill
Preprint. Under review. Patent pending.

\newpage

\section{Introduction}
\textbf{Time Series Forecasting}\\
Time series forecasting plays a critical role in decision-making processes across diverse fields including meteorology, traffic, electricity, economics, finance, and so on. Especially, predicting returns on financial instruments is a challenging problem.\\
\\
\textbf{Time Series Foundation Models}\\
Traditionally, time series forecasting has primarily relied on statistical methods such as the ARIMA model. Inspired by the performance improvements of Large Language Models (LLMs), recent research has proposed time series foundation models pretrained on diverse real-world time series data and capable of addressing various time series prediction tasks. Notable examples include Chronos [1] and Mamba4Cast [2]. The training data for these foundation models significantly differs from language models in terms of both content and acquisition conditions, often containing noise and missing values. Consequently, the quality and design of training data have become crucial problems in building foundation models with sufficient accuracy. Performance-wise, some studies have reported that these foundation models still cannot match the prediction performance of specific time series neural networks trained for financial time series [3].\\
\\
\textbf{Model Training Using Chaotic Time Series}\\
An alternative approach involves utilizing chaotic time series as training data. This methodology is based on the recognition that real-world time series exhibit chaotic properties—characterized by sensitive dependence on initial conditions and nonlinearity. A number of well-known chaotic differential equations exist, including the Lorenz model, which has inspired techniques for artificially generating simple or complex chaotic time series using such equations to create diverse datasets for training.\\
\\
Reference [4] presents a model pretrained using a dataset generated from novel synthetic chaotic time series created through evolutionary algorithms. This model achieves high-accuracy predictions for unseen real-world chaotic sequences—including chemical reactions, fluid dynamics, and brain activity—without requiring retraining. Reference [5] proposes a multivariate RNN-based Mixture-of-Experts model pretrained on 34 types of chaotic time series. This model also demonstrates the ability to predict unseen real-world time series data—such as electric transformer temperatures, traffic volumes, temperature readings, and MRI waveforms—without retraining, achieving superior accuracy compared to conventional time series foundation models like Chronos, Mamba4Cast, and Tiny Time Mixers.\\
\\
\textbf{Are Financial Time Series Chaotic?}\\
Generally, chaotic time series are sequences where future states are determined by deterministic laws of underlying dynamical systems based on current or past conditions. Major subjects in time series forecasting—including weather patterns, traffic flow, and electricity generation—as well as fields like chemical reactions, disease outbreaks, biological responses, and climate change, have all been found to exhibit chaotic behavior. These data primarily follow physical laws, control theory, and statistical models (such as heat conduction, transport theory, atmospheric dynamics, and biological signal control). They also typically feature relatively limited external disturbances, temporal stability with causal relationships, stationarity, periodicity, and minimal noise.\\
\\
However, the question of whether financial time series—particularly stock and currency return data—exhibits chaotic behavior has been the subject of ongoing debate. Reference [6] conducted close return tests on daily data for stock market indices across major countries, confirming that while stock price data shows significant nonlinearity, it does not exhibit chaotic properties. This perspective maintains that financial time series data—including stock prices—are primarily explained by stochastic factors such as nonlinearity, conditional variance, and regime shifts, with limited robust evidence of low-dimensional deterministic chaos [7][8].\\
\\
On the other hand, some studies have reported discoveries of chaotic properties in financial time series. Reference [9] argues that the difficulty in observing chaotic properties in financial time series stems from information loss during the sampling process, suggesting that deterministic elements may exist in auction mechanisms and microstructure. Using high-frequency tick-level time series data for 14 major currency pairs in the foreign exchange market, this study found evidence for chaotic signals. Reference [10] examined 12 currency pairs in the foreign exchange market and confirmed signs of deterministic chaos in all pairs, even in daily data.\\
\\
\textbf{Correspondence Between Chaotic Time Series and Financial Time Series}\\
Since financial time series contain significant noise and low-dimensional state space is rarely observed, inferring underlying dynamical phenomena from data alone is almost impossible. However, simple models for volume and price fluctuations can sometimes be derived, and their results may correspond to equations of chaotic time series. Reference [11] developed a simplified model of the mutual relationship between price and volume fluctuations with expectation formation of market participants, leads to economic equations. From these equations, it was confirmed that the simple harmonic oscillations exist in price fluctuations, and that these equations can assume the form of the Lorenz model of chaotic time series, as demonstrated mathematically.\\
\\
Furthermore, certain characteristics of various chaotic time series have been utilized in generating financial data and simulations. The Henon map—a type of chaotic time series—is an iterative model that can be applied directly to discrete data such as daily or minute-by-minute data. Its highly sensitive dependence on initial conditions—where small differences lead to dramatically divergent outcomes—resembles the amplification of market shocks, making it sometimes used for generating sequences of stock returns or volatility series [12]. The Mackey-Glass equation, another chaotic time series, includes delay terms that make it particularly suitable for modeling decision-making delays and information propagation lags among market participants. Its ability to easily simulate volatility clustering and self-correlation structures has led to its use in generating training data for predictive models of stock returns [13].\\
\\
\textbf{Approach to Financial Time Series Forecasting in this paper}\\
Despite these ongoing debates, this paper considers financial time series to possess chaotic properties. However, as explained above, when comparing financial time series to other real-world time series with known dynamical systems and to various theoretical chaotic time series, the behavior clearly differs (Fig. 1). With agreement to the perspective presented in Reference [9], we propose that chaotic properties exist at the microstructure level—even at the temporal scale of individual trading participants—and that financial time series are formed through discrete observations. We generate data simulating financial time series by resampling chaotic time series generated at typical integration time intervals, then use this as training data. Traditional applications of predicting chaotic time series use Seq-to-Seq methods to forecast long-term sequences up to 512 integration time units, while this paper adopts one-period-ahead prediction in the predictive horizon and uses resampled sequences with the same interval as the context for prediction (Fig. 2).\\
\\
Even if chaotic properties exist at extremely short microstructure time scales, observing its chaotic sequence remains difficult. Adopting resampled sequences as context can facilitate their application to real-world financial time series data such as minute-by-minute or hourly data. When predicting financial time series, particularly returns, extending the predictive horizon to obtain larger expectation profit is demanding. While expanding the resampling interval to extend predictive horizons reduces autocorrelation and increases prediction difficulty, we leverage the advantage of infinitely generated chaotic time series to perform data augmentation and improve prediction performance.\\
\\
\textbf{Large-Scale Pretraining Using Chaotic Time Series}\\
We adopt the Lorenz model for the chaotic time series equations used in training. Assuming that extremely short-term microstructure time scales possess chaotic properties, we anticipate that the interactions between variables in the Lorenz model may simulate relations between price, volume, and order flow, as demonstrated in Reference [11]. We considered identifying Lorenz model parameters from specific asset's data, but due to the significant noise components in financial time series and the uncertainty in parameter reliability, as well as our intention to apply the model to multiple assets, we instead adopted representative parameter values for the Lorenz model and generated training samples by introducing perturbations to these parameters.\\
\\
We conducted large-scale pretraining with 10 billion training samples for each predictive horizon by expanding the resampling interval. The model architecture employs a Decoder-type Transformer, using Attention Masks to learn Seq-to-Seq predictions for each element in the context sequence. While chaotic time series typically become increasingly difficult to forecast as the predictive horizon extends, this paper has observed a scaling law-like phenomenon that we can achieve predictive performance at a certain level with extended predictive horizons for chaotic time series by increasing the number of training samples exponentially. Additionally, during the training of long predictive horizon, we observed gradual formation of the state space (attractor) as training samples increased, indicating progressive learning progress.\\
\\
\textbf{Predicting Real-World Financial Time Series}\\
We create a test dataset from actual financial instrument data and perform zero-shot prediction using the pretrained model. Here, we use readily accessible Bitcoin trade data, aggregating it at 5-second, 10-second, 15-second, 20-second, 25-second, 30-second, and 60-second timeframes to create sequences of price, volume, and order flow, then scale these to train data and finally create the test dataset. Using these test data at each timeframe, we perform prediction using a pretrained model with a predictive horizon of 100, 300, 500, 700, and 1000 unit time. Testing a simple trading strategy based on these predictions confirmed superior performance compared to autocorrelation models.

\newpage


\newpage

\begin{figure}[H]
  \centering
  \begin{subfigure}{0.8\linewidth}
    \centering
    \includegraphics[width=\linewidth]{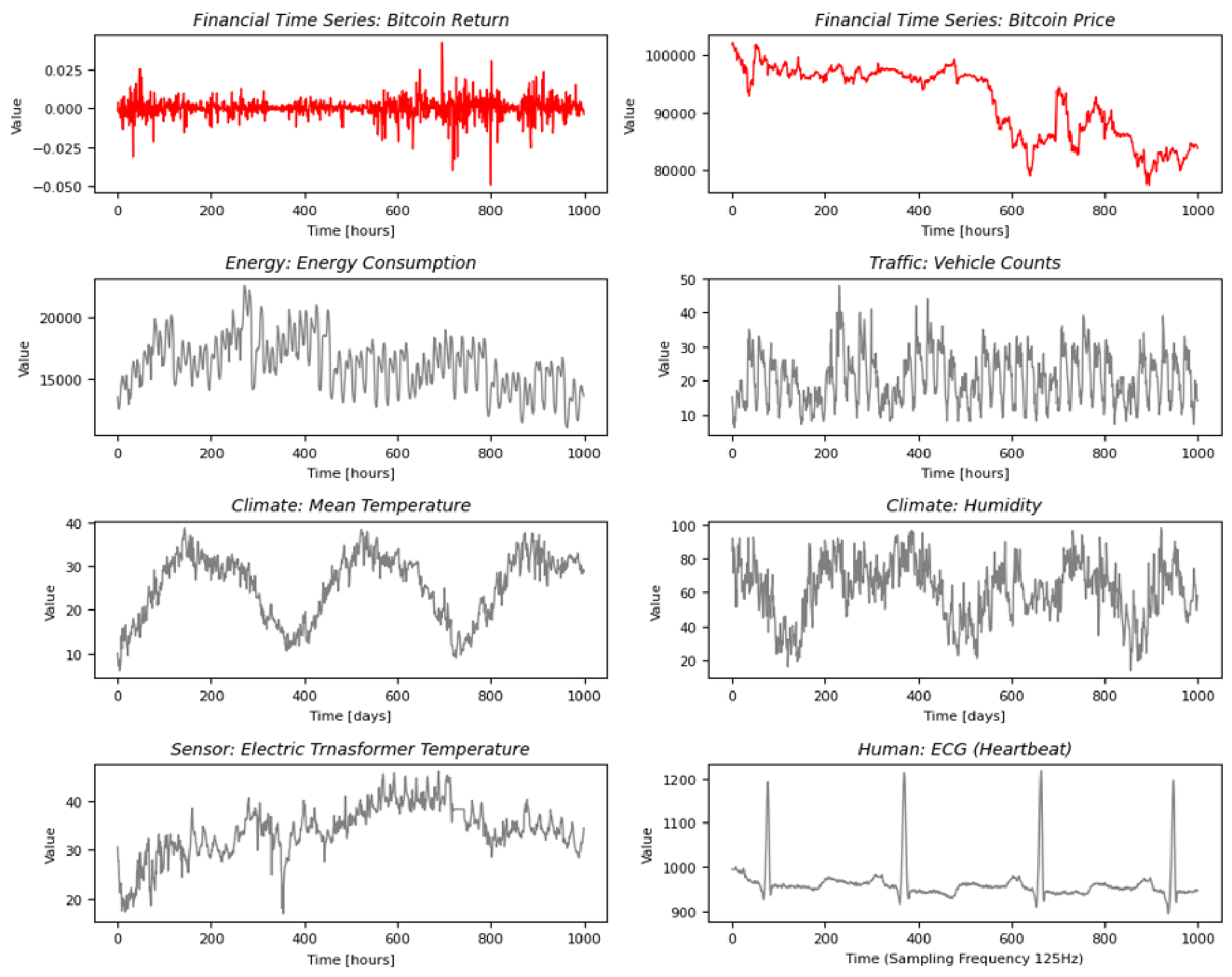}
    \subcaption*{(a)}
    \label{fig1-a}
  \end{subfigure}
  \vspace{1em}
  \begin{subfigure}{0.8\linewidth}
    \centering
    \includegraphics[width=\linewidth]{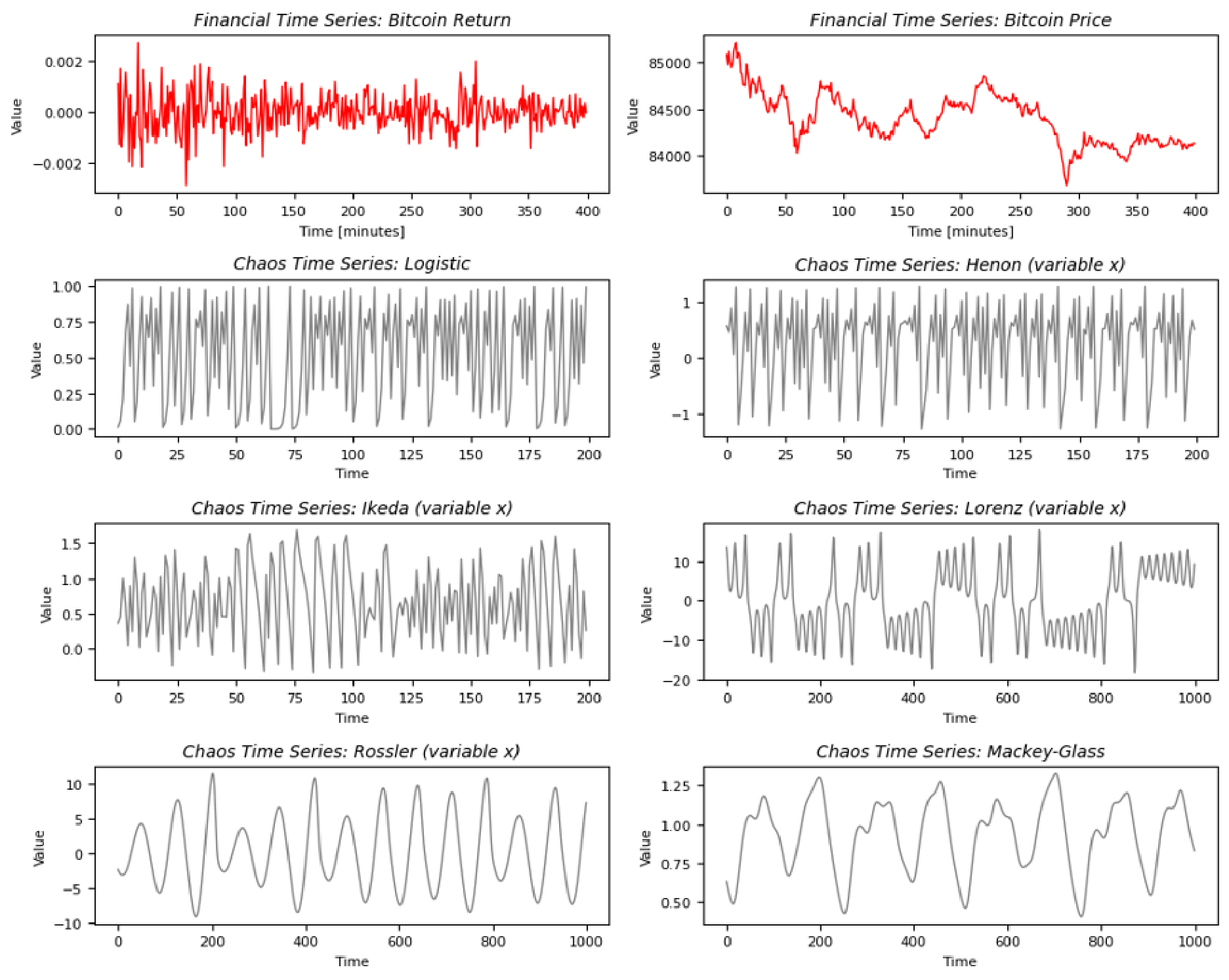}
    \subcaption*{(b)}
    \label{fig1-b}
  \end{subfigure}
  \caption{a) Financial time series versus various other real-world time series waveforms. b) Financial time series versus various other chaotic time series waveforms.}
  \label{fig1}
\end{figure}

\begin{figure}[H]
  \centering
  \includegraphics[width=0.8\textwidth]{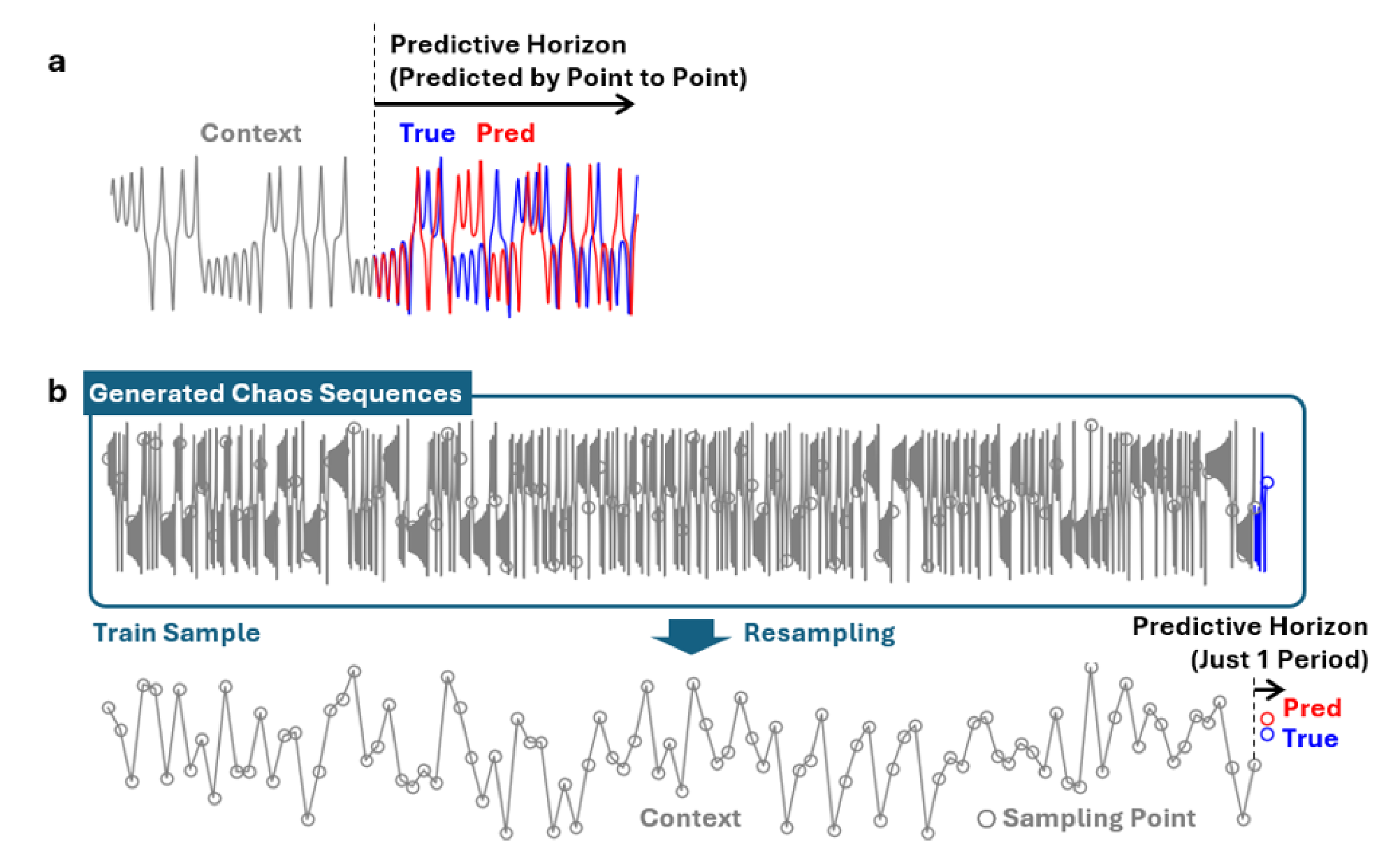}
  \caption{a) Conventional Seq-to-Seq method for predicting chaotic time series. b) The methodology for generating training samples and prediction approach employed in this paper.}
  \label{fig2}
\end{figure}

\section{Method}
\subsection{Dataset}
As discussed in previous chapters, this paper generates financial time series data artificially by resampling chaos time series created with a commonly used integration time (unit time) for training samples. For creating the chaos time series, we use the Lorenz model to simulate price, volume, and order flow, as outlined in reference [11].

\begin{equation}
\begin{cases}
\dot{x} = \sigma\,(y - x),\\
\dot{y} = x(\rho - z) - y,\\
\dot{z} = x y - \beta z,
\end{cases}
\label{lorenz}
\end{equation}

The parameters and initial values for the Lorenz model are determined by randomly sampling from a uniform distribution within the ranges of $\sigma \in [9,11],\;\rho \in [26,30],\;\beta \in [2.3,3.1],\; x,y,z \in [0.18,0.22]$, with representative values set as $\sigma$=10, $\rho$=28, $\beta$=3/8, and (x0, y0, z0)=(0.2, 0.2, 0.2). The integration time $\Delta$t is set to 0.01.\\
\\
The context length for model input is 512, and the resampling interval for determining predictive horizons is set to [100, 200, 300, 400, 500, 700, 1000]. For example, when the resampling interval is set to 1000, the model generates 513×1000 sequences of chaos time series per integration time $\Delta$t (plus a warm-up period), from which 513 simulated financial time series are obtained by resampling every 1000 intervals (512 context + 1 prediction period). As in standard LLM training, loss is computed by seq-to-seq for each context element with one-period-ahead prediction, meaning this single sequence constitutes 512 training samples. Since random seed numbers are regenerated for each sequence generation, there is no same contexts during the model training processes. To achieve a training sample count of 10 billion, generating sequences with a resampling interval of 1000 would require 1e13 total time series generation.\\
\\
For each resampling interval, we analyze various characteristics of the generated sequences, as shown in Fig.3. We observe that while autocorrelation decreases with increasing resampling interval, the shape of the state space (attractor) remains same formation. Assuming real-world financial time series represent discrete observations of chaotic sequences, the difficulty in detecting state space is not due to information loss during discrete observations, but rather stems from their high dimensionality or observational noise.

\begin{figure}[H]
  \centering
  \begin{subfigure}{0.8\linewidth}
    \centering
    \includegraphics[width=\linewidth]{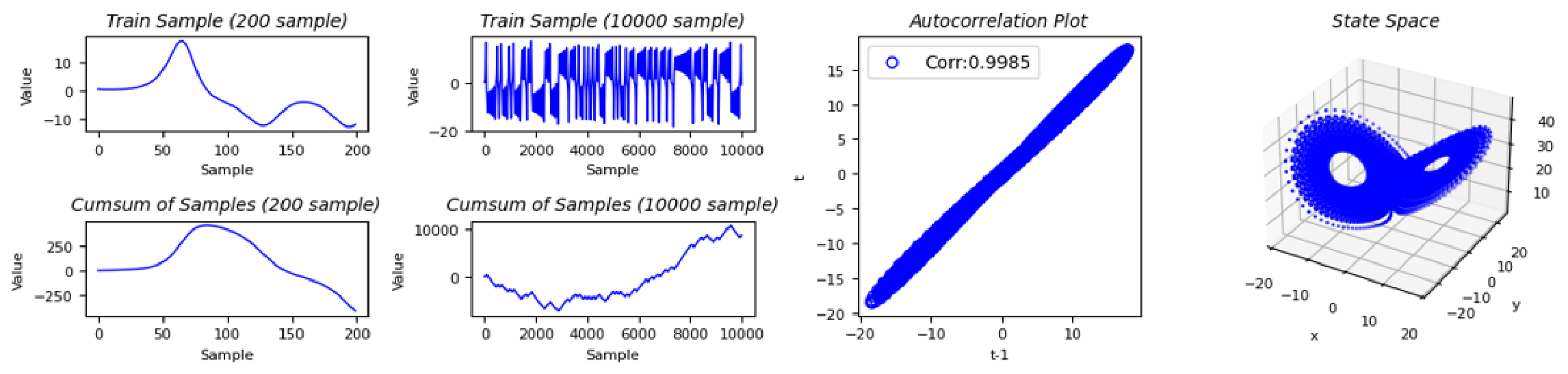}
    \subcaption*{(a)}
    \label{fig3-a}
  \end{subfigure}
  \vspace{1em}
  \begin{subfigure}{0.8\linewidth}
    \centering
    \includegraphics[width=\linewidth]{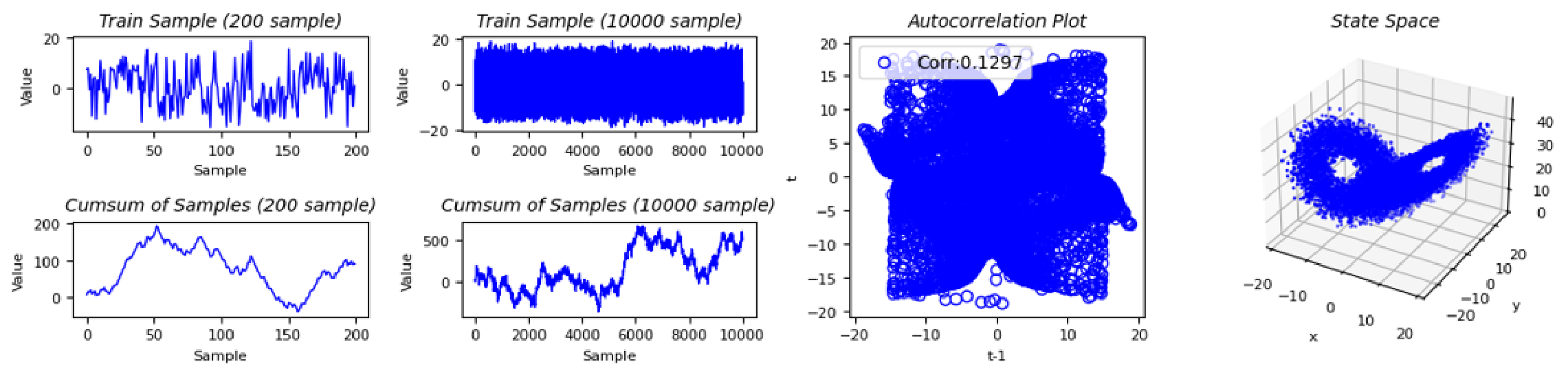}
    \subcaption*{(b)}
    \label{fig3-b}
  \end{subfigure}
  \vspace{1em}
  \begin{subfigure}{0.8\linewidth}
    \centering
    \includegraphics[width=\linewidth]{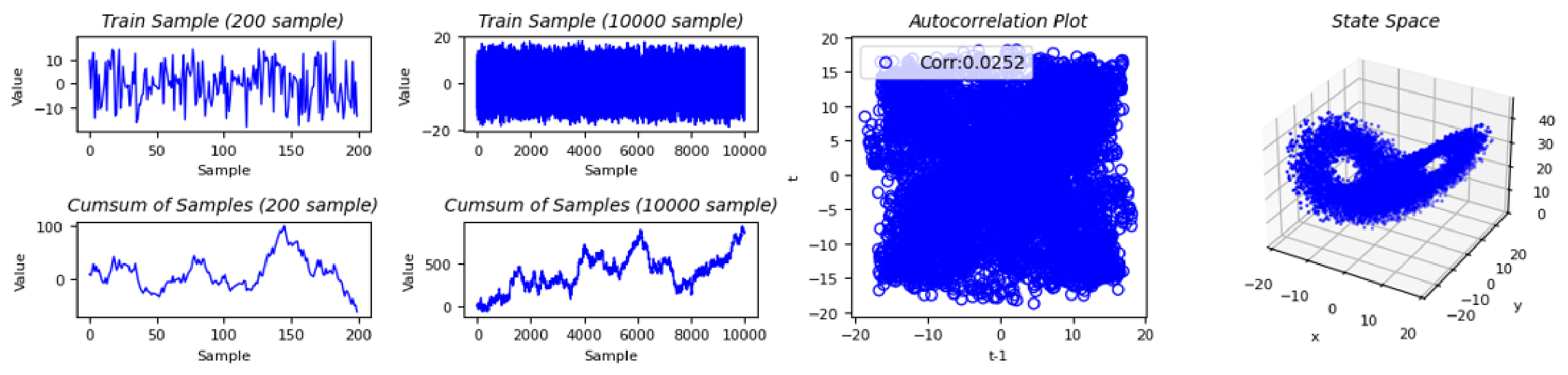}
    \subcaption*{(c)}
    \label{fig3-c}
  \end{subfigure}
  \vspace{1em}
  \begin{subfigure}{0.8\linewidth}
    \centering
    \includegraphics[width=\linewidth]{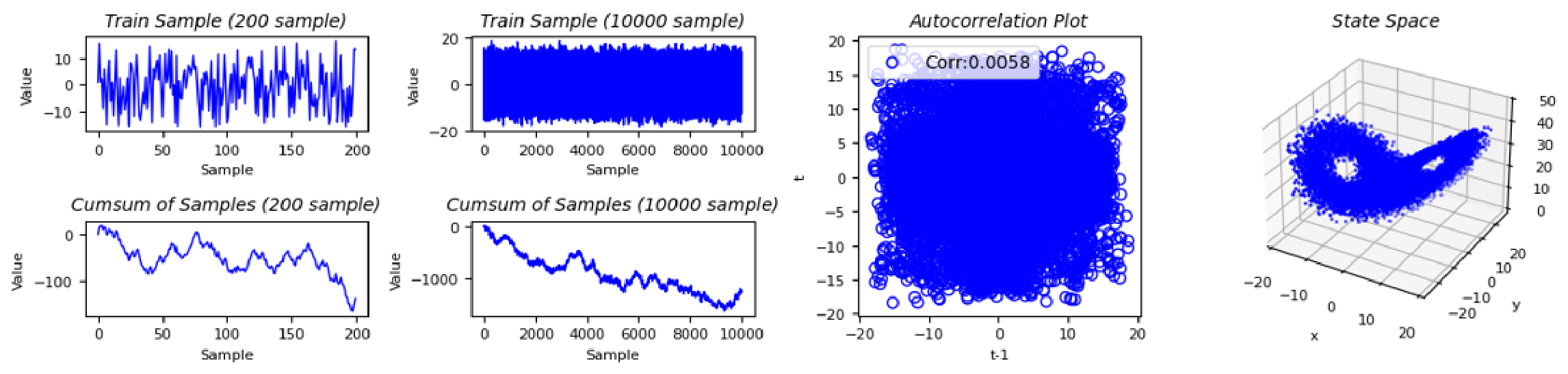}
    \subcaption*{(d)}
    \label{fig3-d}
  \end{subfigure}
    \caption{(first half)}
  \label{fig3}
\end{figure}

\begin{figure}[H]
  \ContinuedFloat
  \centering
  \begin{subfigure}{0.8\linewidth}
    \centering
    \includegraphics[width=\linewidth]{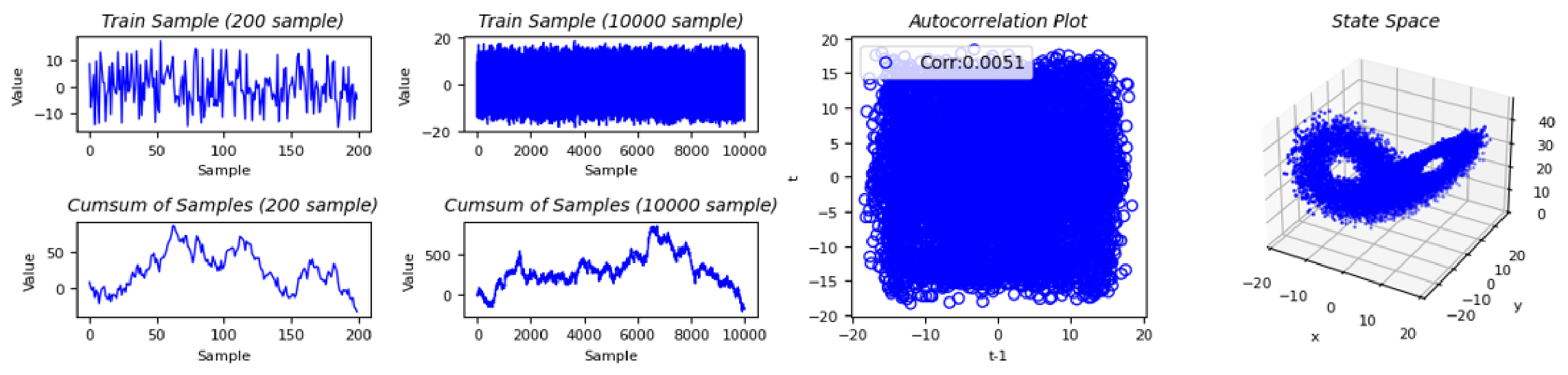}
    \subcaption*{(e)}
    \label{fig3-e}
  \end{subfigure}
  \vspace{1em}
  \begin{subfigure}{0.8\linewidth}
    \centering
    \includegraphics[width=\linewidth]{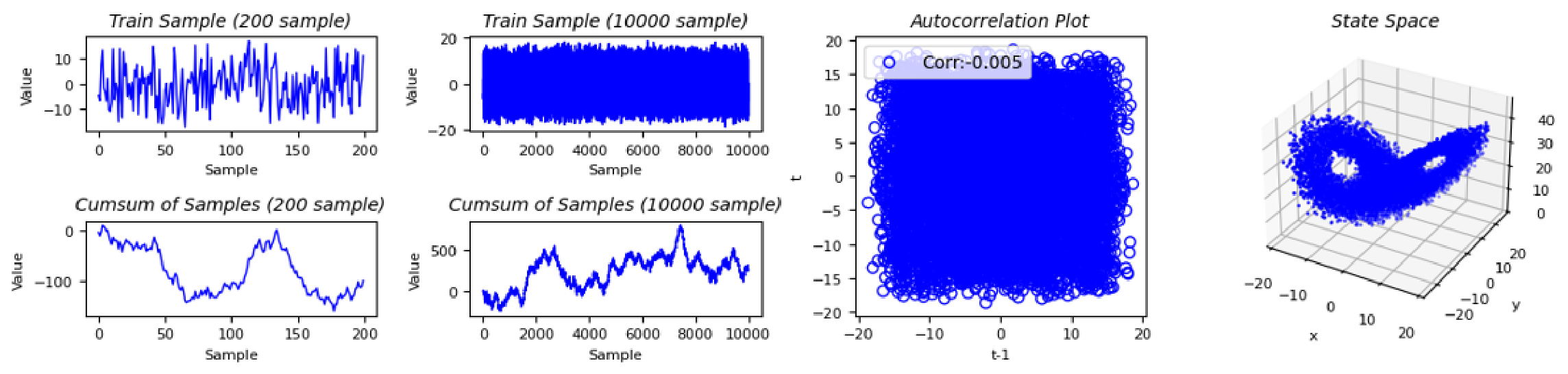}
    \subcaption*{(f)}
    \label{fig3-f}
  \end{subfigure}
  \caption{10,000 sequences of representative training sample waveforms across different resampling intervals. a)–f) Each figure corresponds to resampling intervals of 1 (no resampling), 100, 300, 500, 700, and 1000. In each figure, from left to right: training sample 200's values of x (upper row: raw sequence, lower row: cumulative sum sequence); training sample 10,000's values of x (upper row: raw sequence, lower row: cumulative sum sequence); autocorrelation between the current and one-period-earlier training samples; and state space (attractor).}
  \label{fig3}
\end{figure}

\subsection{Model Architecture}
In this study, we adopt a decoder-type Transformer architecture for time series forecasting. Originally proposed for natural language processing tasks [14], the Transformer has been widely applied to time series modeling due to its ability to capture long-range dependencies through the self-attention mechanism [15][16].\\
\\
Given an input sequence $\mathbf{x} = \{x_1, x_2, \ldots, x_T\}$, each scalar observation is mapped into an embedding vector $\mathbf{z}_t \in \mathbb{R}^d$. Positional encodings are added to preserve the sequential order:
\begin{equation}
\mathbf{h}_t^{(0)} = \mathbf{E}(x_t) + \mathbf{p}_t, \quad t = 1, \ldots, T,
\end{equation}
where $\mathbf{E}$ denotes the learnable embedding mapping and $\mathbf{p}_t$ the positional encoding.\\
\\
\textbf{Self-Attention with Causal Mask}\\
Each decoder layer applies masked self-attention. The query, key, and value matrices are defined as
\begin{equation}
Q = H W_Q, \quad K = H W_K, \quad V = H W_V,
\end{equation}
where $H \in \mathbb{R}^{T \times d}$ is the input representation and $W_Q, W_K, W_V \in \mathbb{R}^{d \times d_k}$ are learnable projection matrices.  
The self-attention output is given by
\begin{equation}
\text{Attention}(Q,K,V) = \text{softmax}\!\left(\frac{QK^\top}{\sqrt{d_k}} + M\right)V,
\end{equation}
where $M$ denotes the causal mask, ensuring that position $i$ cannot attend to future positions $j>i$.\\
\\
\textbf{Layer Composition}\\
Each decoder layer consists of (1) masked self-attention, (2) residual connections with layer normalization, and (3) a position-wise feed-forward network (FFN), defined as
\begin{equation}
\text{FFN}(h) = \sigma(h W_1 + b_1) W_2 + b_2,
\end{equation}
where $\sigma$ is a non-linear activation function, typically ReLU or GELU. Stacking $L$ such layers yields the final hidden representations $\mathbf{H}^{(L)}$.\\
\\
\textbf{Output Layer}\\
For forecasting, the model takes the most recent sequence of length $T$ and autoregressively generates future values. The decoder output is mapped to scalar predictions through a linear projection:
\begin{equation}
\hat{x}_{t+1} = \mathbf{h}_t^{(L)} \mathbf{W}_o + b_o,
\end{equation}
where $\mathbf{W}_o \in \mathbb{R}^{d \times 1}$ and $b_o$ are learnable parameters. By combining causal self-attention with stacked decoder layers, the Transformer can effectively capture both local and long-term temporal dependencies, making it well-suited for time series forecasting.\\
\\
To verify prediction performance across different model sizes, we train three model variants with parameter counts of 0.1M, 1M, and 10M. Table 2.1 lists the specifications of these models, including nparams (approximate parameter count), nlayers (number of layers), dmodel (dimension per layer), and dhead (number of attention heads). All models use a context length of 512.

\begin{table}[H]
\centering
\caption{Model parameters.}
\begin{tabular}{lccccccc}
\toprule
Model Name & $n_{\text{params}}$ & $n_{\text{layers}}$ & $d_{\text{model}}$ & $n_{\text{heads}}$ & Batch Size & Learning Rate \\
\midrule
Model 0.1M & 0.1M & 4  & 64  & 4  & 60 & $1.0\times 10^{-3}$ \\
Model 1M   & 1M   & 10 & 128 & 8  & 60 & $1.0\times 10^{-3}$ \\
Model 10M  & 10M  & 12 & 384 & 24 & 40 & $1.0\times 10^{-3}$ \\
\bottomrule
\end{tabular}
\label{tab:model_config}
\end{table}

\subsection{Training Process}
The training process was adapted from the training conditions of GPT-3 [17]. The optimizer is Adam, with parameters set to $\beta_1$=0.9, $\beta_2$=0.95, and 0.1 weight decay applied for regularization. Gradient global norms are clipped at 1.0. The learning rate is warmed up over 6\% of the total training samples before gradually decreasing according to a cosine decay schedule. Batch size was fixed during training. To prevent overfitting, the entire 10 billion training samples were trained over a single epoch. Loss computation uses MSE loss, calculated across each sequence's three dimensions from the Lorenz model and summed together. Data is generated batch-by-batch, backpropagated, and used to update parameters. The model was not parallelized and was trained on an A100 GPU.

\subsection{Test Data}
We conduct zero-shot prediction on real-world financial time series without retraining to explore the applicability of pretrained models with chaos time series to actual time series data. For real-world financial time series, we use readily accessible Bitcoin trade data. Specifically, we utilize trade data from the BTCUSDT perpetual futures contract on the cryptocurrency exchange Bybit from January to February 2025. To verify which predictive horizon is effective for several real-world timeframe, we reaggregate the trade data at timeframes of 5sec, 10sec, 15sec, 20sec, 25sec, 30sec, and 60sec. To map real-world financial time series to the Lorenz model's variables x, y, and z, we propose the following correspondences based on reference [11].

\begin{equation}
\begin{cases}
\dot{x} = \sigma\,(y - x),\\
\dot{y} = x(\rho - z) - y,\\
\dot{z} = x y - \beta z,
\end{cases}
\tag{\ref{lorenz}}
\end{equation}

x represents order flow, which also serves as an alternate variable indicating very short-term participant sentiment. y denotes price change rate, with predicted values of y serving as indicators for trading decision-making. z represents volume, indicating the overheating levels of market. This establishes the following economic interpretations for each equation in the Lorenz model:

\begin{itemize}
  \item Equation 1 shows how order flow x follows and increases in response to price change y, while simultaneously experiencing self-dampening effects. It describes a dynamic where order flow (sentiment) responds to short-term price fluctuations while also self-correcting. $\sigma$ is the reaction speed parameter, representing how sensitive investor sentiment and order flow are to price movements. A large $\sigma$ indicates a hypersensitive market where sentiment immediately reacts to price changes, while a small $\sigma$ suggests a less responsive market where sentiment tends to diverge from price movements.
  \item Equation 2 shows how price change y is driven by order flow x, but with the effect being suppressed and reversed when volume z exceeds a certain threshold (indicating market overheating). It describes a dynamics where prices fluctuate sentiment-driven, but under excessive volume conditions, price increases (or decreases) become inhibited and reversed. $\rho$ is the sentiment amplification parameter, determining both the magnitude of the positive effect of order flow on price change rate and the threshold at which market behavior reverses in relation to volume. A large $\rho$ indicates persistent price push-up (or push-down) effects from sentiment, with significant reversals occurring when volume spikes sharply. A small $\rho$ suggests limited order flow effects, with prices remaining stable and quick reversals occurring with minor volume surges.
  \item Equation 3 shows how volume z increases when order flow x and price change y strengthen in the same direction. However, market expectations and enthusiasm cool, causing volume to self-dampen over time. It describes a dynamic where price-flow synergy leads to volume explosions, which then gradually subside through self-dampening. $\beta$ is the volume decay parameter determining how quickly volume naturally decays or converges. It represents the speed at which market overheating dissipates and the duration of liquidity shock persistence. A large $\beta$ means volume peaks quickly dissipate, while a small $\beta$ indicates prolonged volume buildup and sustained overheating.
\end{itemize}

Figure 4 shows state space plots of variables x, y, and z generated from the Lorenz model using resampling intervals of 1000, alongside the corresponding order flow, price change rate, and volume from real-world financial time series. Comparative observation of the state spaces of chaotic time series and financial time series reveals visually similar distributions in XY, YZ, and ZX cross-sections, though in three-dimensional state space, no distinct features resembling the attractor of chaotic time series are observed in financial time series. This is likely due to the high dimensionality of financial time series' state space and observational noise. Finally, we create 100,000 sequences of each of the three-dimensional series for each timeframe from 5sec to 60sec, calculate the scaling coefficients with the initial 10,000 sequences, then apply these to the remaining 90,000 sequences as test data. While prediction uses the same 512 context length as training, prediction is performed only for one period ahead within the context, with context sliding one period at a time to predict at all time points.

\begin{figure}[H]
  \centering
  \includegraphics[width=0.8\textwidth]{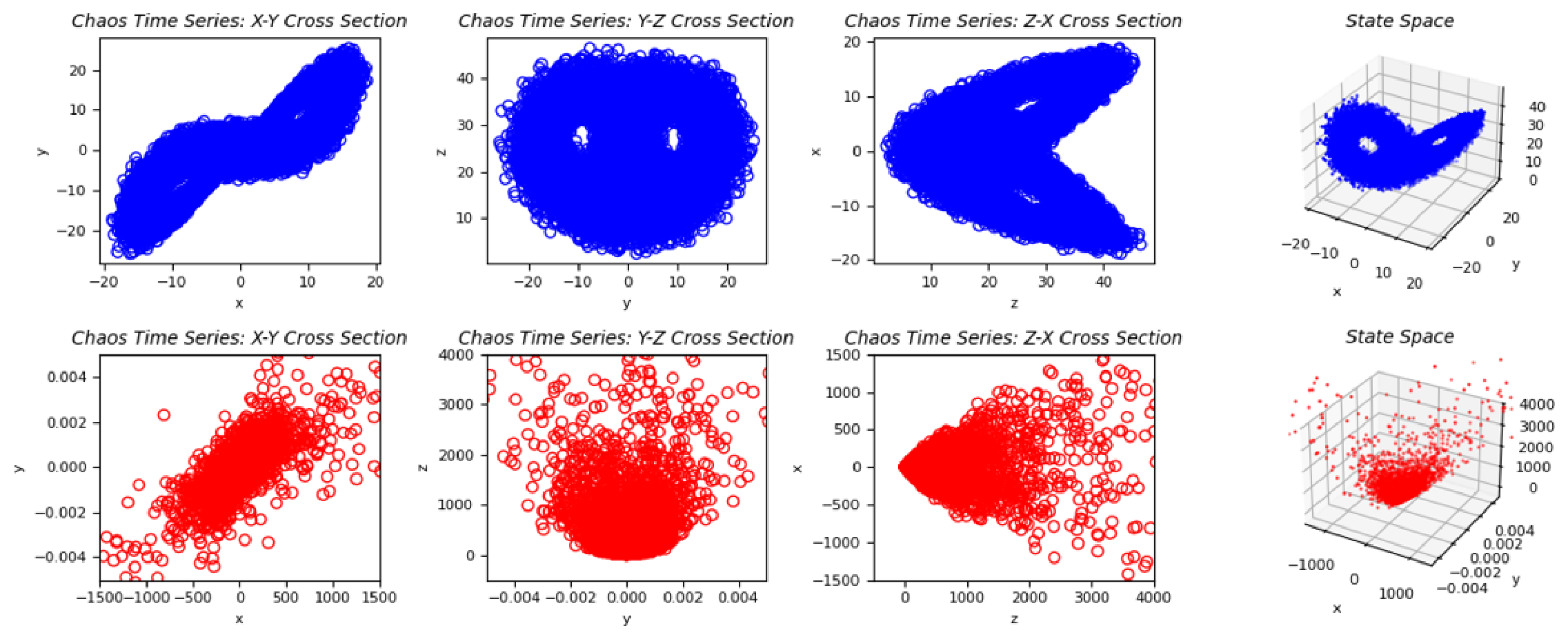}
  \caption{State spaces (attractors) for each time series. Top row: chaotic time series generated using the Lorenz model with a resampling interval of 1,000. Bottom row: sequences representing order flow, price change rate, and volume from actual financial time series, with x, y, and z respectively (timeframe: 30 seconds). Left to right: XY cross-section, YZ cross-section, ZX cross-section, and 3D spatial representation.}
  \label{fig4}
\end{figure}

\section{Results}
\subsection{Learning Results for Chaotic Time Series}
Figure 5 shows Seq-to-Seq prediction results for chaotic time series generated using seeds not included in the training set. The model parameters used are 1M. Here, we have selected examples with predictive horizons of 500, 700, and 1000 unit time from top to bottom. The time series waveforms shown on the left are expanded views of 512 samples each, while the right side displays the full sequence of all validation data. The three-dimensional graphs on the right represent the attractor shapes in the one-step-ahead target state space on the left, and the predicted results plotted in the same state space, respectively.\\
\\
For the 500-unit time prediction results, the model accurately predicts each variable of the Lorenz model, and the attractor shape is successfully reconstructed in the state space. For the 700-unit time prediction results, while variable z shows reasonable prediction performance, variables x and y often produce predictions near the zero level. However, occasional pulse-like prediction outputs are observed. This indicates that under relatively predictable conditions within complex contexts, the model can generate such predictions and gradually reconstruct the attractor. For the 1000-unit time prediction results, the frequency of pulse-like predictions decreases further. Nevertheless, we can clearly observe the model gradually reconstructing the attractor by dispersing predictions across the spatial domain.

\begin{figure}[H]
  \centering
  \begin{subfigure}{0.8\linewidth}
    \centering
    \includegraphics[width=\linewidth]{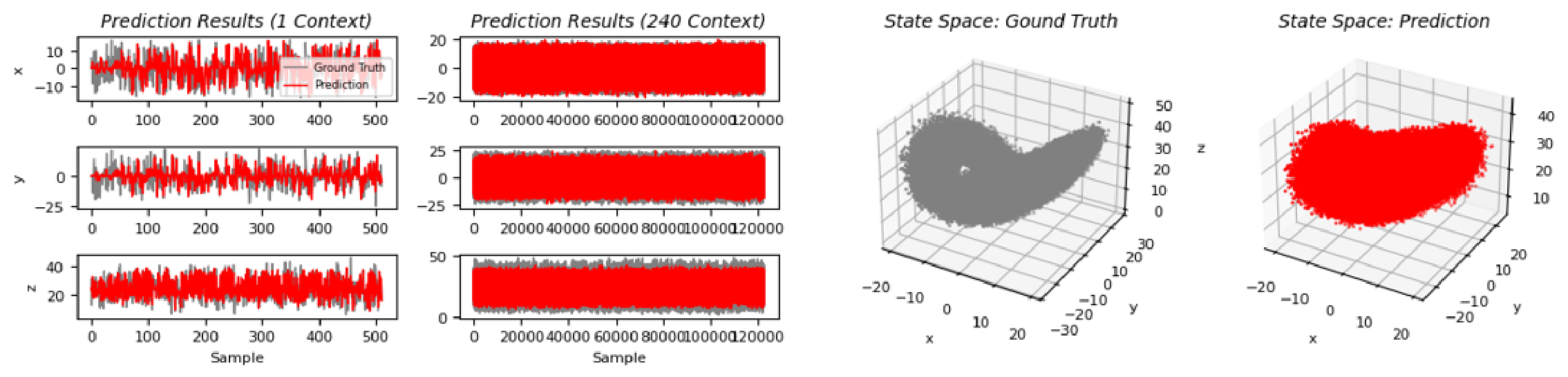}
    \subcaption*{(a)}
    \label{fig5-a}
  \end{subfigure}
  \vspace{1em}
  \begin{subfigure}{0.8\linewidth}
    \centering
    \includegraphics[width=\linewidth]{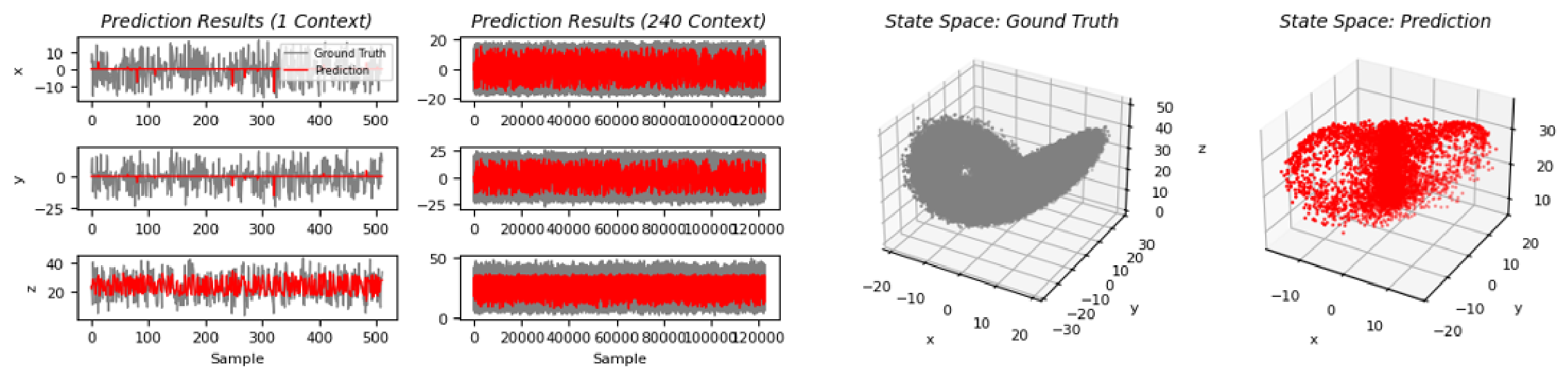}
    \subcaption*{(b)}
    \label{fig5-b}
  \end{subfigure}
  \vspace{1em}
  \begin{subfigure}{0.8\linewidth}
    \centering
    \includegraphics[width=\linewidth]{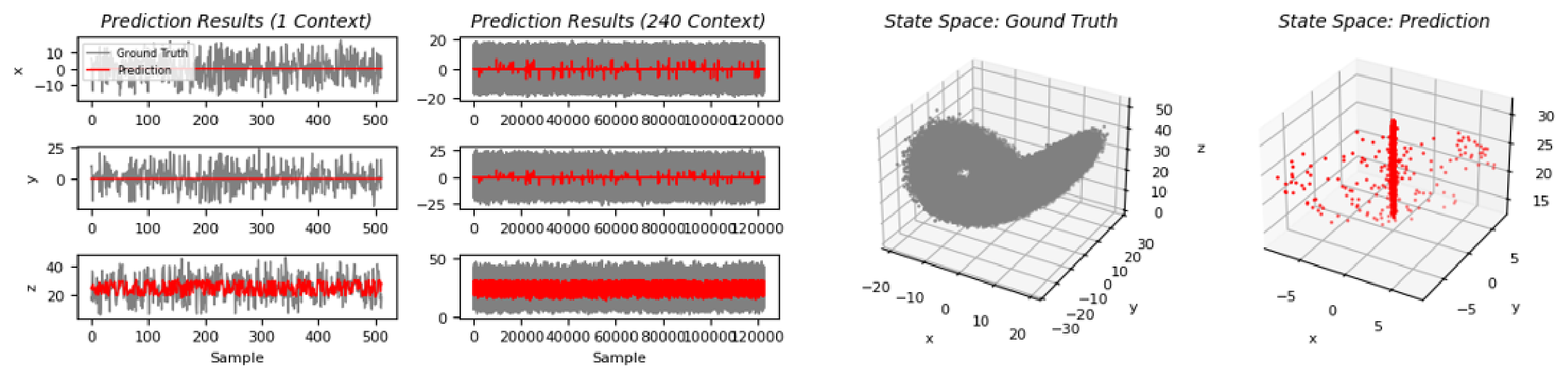}
    \subcaption*{(c)}
    \label{fig5-c}
  \end{subfigure}
  \caption{Prediction results for chaotic time series. a)–c) Each figure corresponds to predictive horizons of 500, 700, and 1,000 unit time. In each figure, from left to right: ranking; single context Seq-to-Seq prediction results (top row: x, y, z values); 240 contexts viewed from a bird's-eye perspective (top row: x, y, z values); state space (attractor) of the context; and state space (attractor) of the prediction output.}
  \label{fig5}
\end{figure}

Figure 6a shows the loss curves for trained samples at each predictive horizon using model parameters of 1M. Figure 6b displays the evolution of the correlation coefficient between the predicted variable y from the Lorenz model (using models under training) and the actual one-step-ahead target. The choice of variable y as the evaluation metric stems from the fact that its predicted values serve as indicators for trading decision-making in previous chapters. In financial return prediction, the correlation coefficient between prediction and target is called the information coefficient and serves as one of the metrics for assessing prediction skill. Shorter predictive horizons correspond to higher series autocorrelation, making predictions easier and enabling high-accuracy predictions with relatively few training samples. For the 700- and 1000-unit time predictive horizons, even after reaching the 10 billion training sample limit, no convergence in loss reduction was observed, suggesting that further training could potentially improve accuracy. During training, we gradually observed an increasing number of samples becoming predictable, leading to the formation of an attractor, indicating promising prospects for future training progress (Figure 7).

\begin{figure}[H]
  \centering
  \begin{subfigure}{0.8\linewidth}
    \centering
    \includegraphics[width=\linewidth]{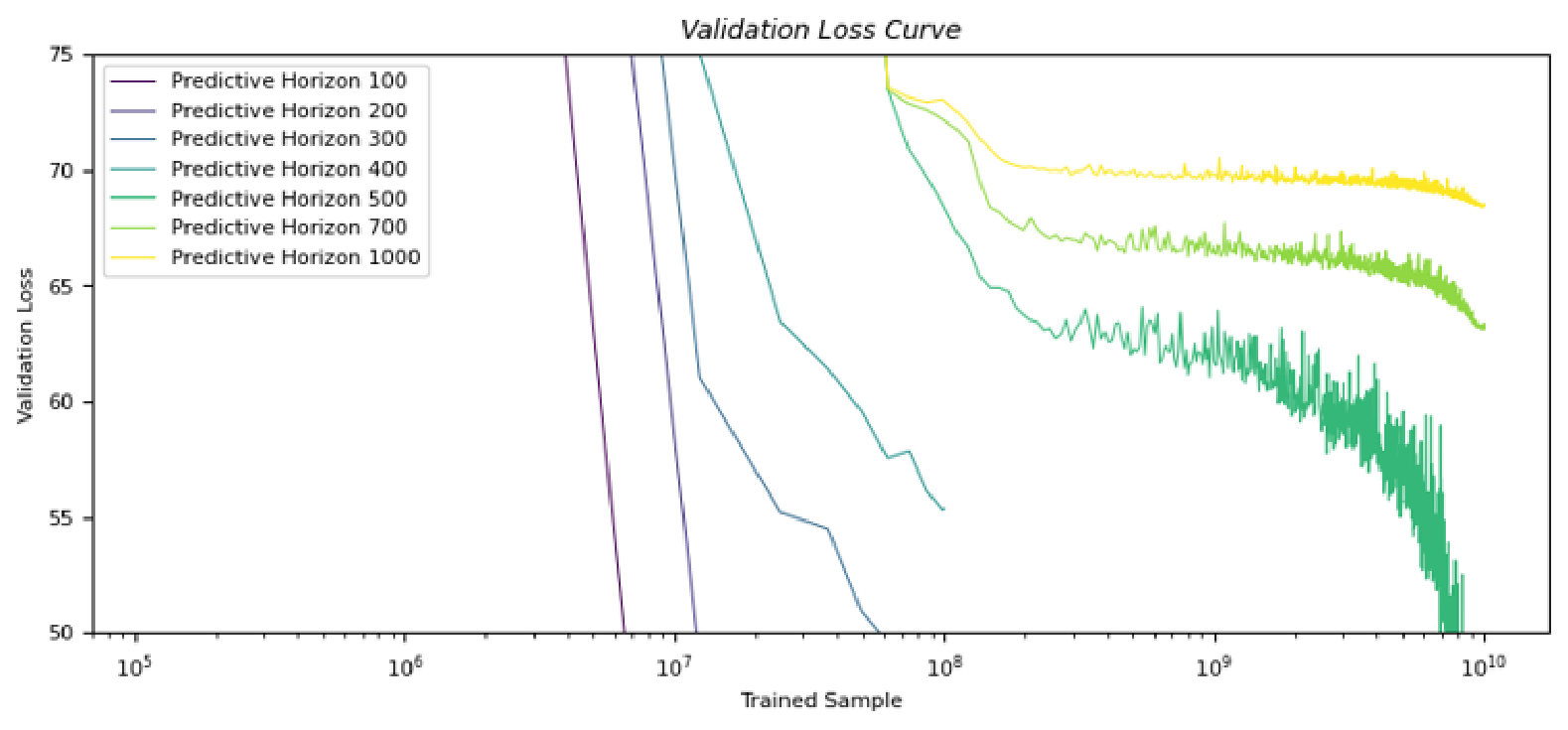}
    \subcaption*{(a)}
    \label{fig6-a}
  \end{subfigure}
  \vspace{1em}
  \begin{subfigure}{0.8\linewidth}
    \centering
    \includegraphics[width=\linewidth]{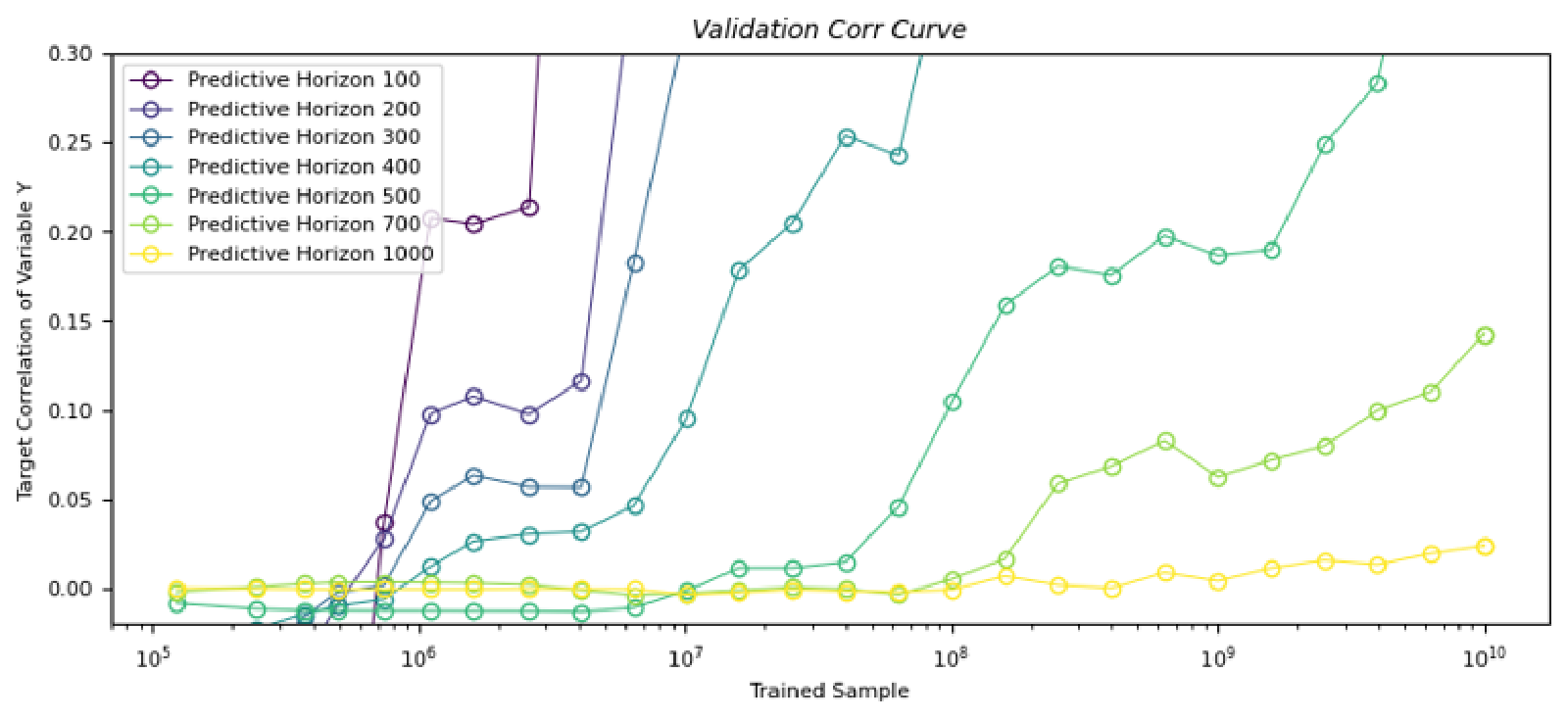}
    \subcaption*{(b)}
    \label{fig6-b}
  \end{subfigure}
  \caption{a) Loss curve as a function of the number of trained samples. b) Correlation coefficient between the prediction output by the model during training and the target value one period ahead (for the Lorenz model's variable y).}
  \label{fig6}
\end{figure}

\begin{figure}[H]
  \centering
  \includegraphics[width=0.8\textwidth]{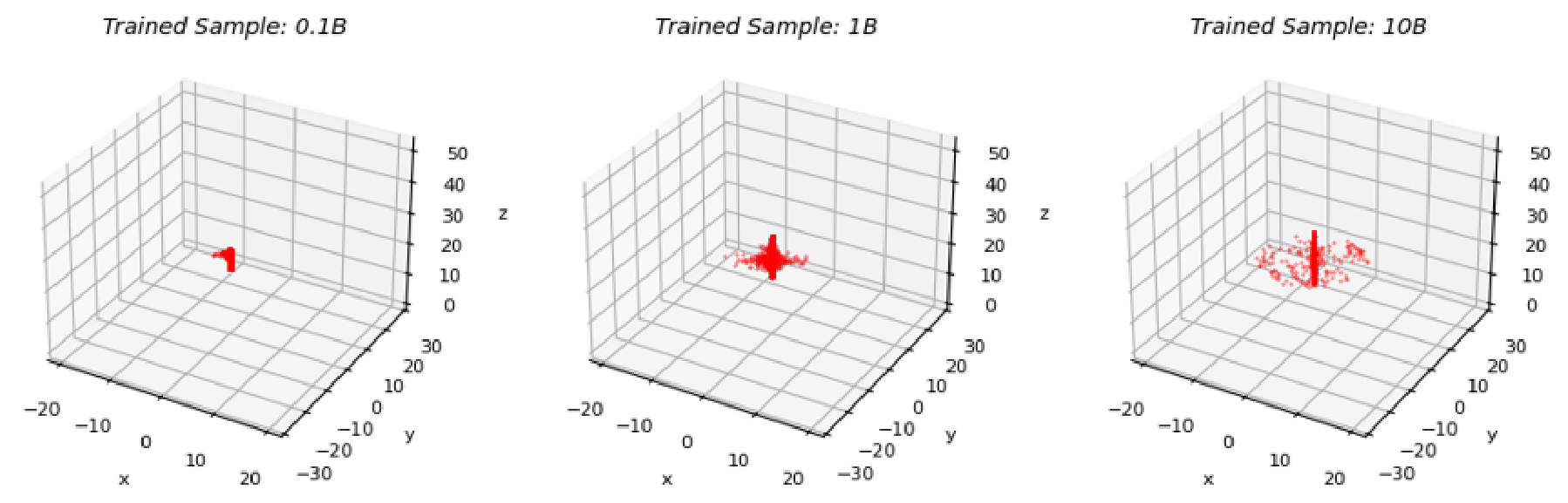}
  \caption{Evolution of state space in prediction outputs (prediction horizon of 1,000 unit time, at completion points of 100 million, 100 million, and 1 billion training samples).}
  \label{fig7}
\end{figure}

Figure 8 shows the loss curves for model parameters of 0.1M, 1M, and 10M, with a predictive horizon of 1000 unit time. While the loss curve for 0.1M parameters is significantly larger compared to others, those for 1M and 10M parameters show nearly identical levels, suggesting that the loss improvement from increasing model parameters may plateau around 1M parameters. One possible explanation for why the model parameters remain remarkably small compared to GPT-3 [14] is that the chaotic time series we handle are just three-dimensional, which is sufficiently smaller than the embedding spaces in language models.

\begin{figure}[H]
  \centering
  \includegraphics[width=0.8\textwidth]{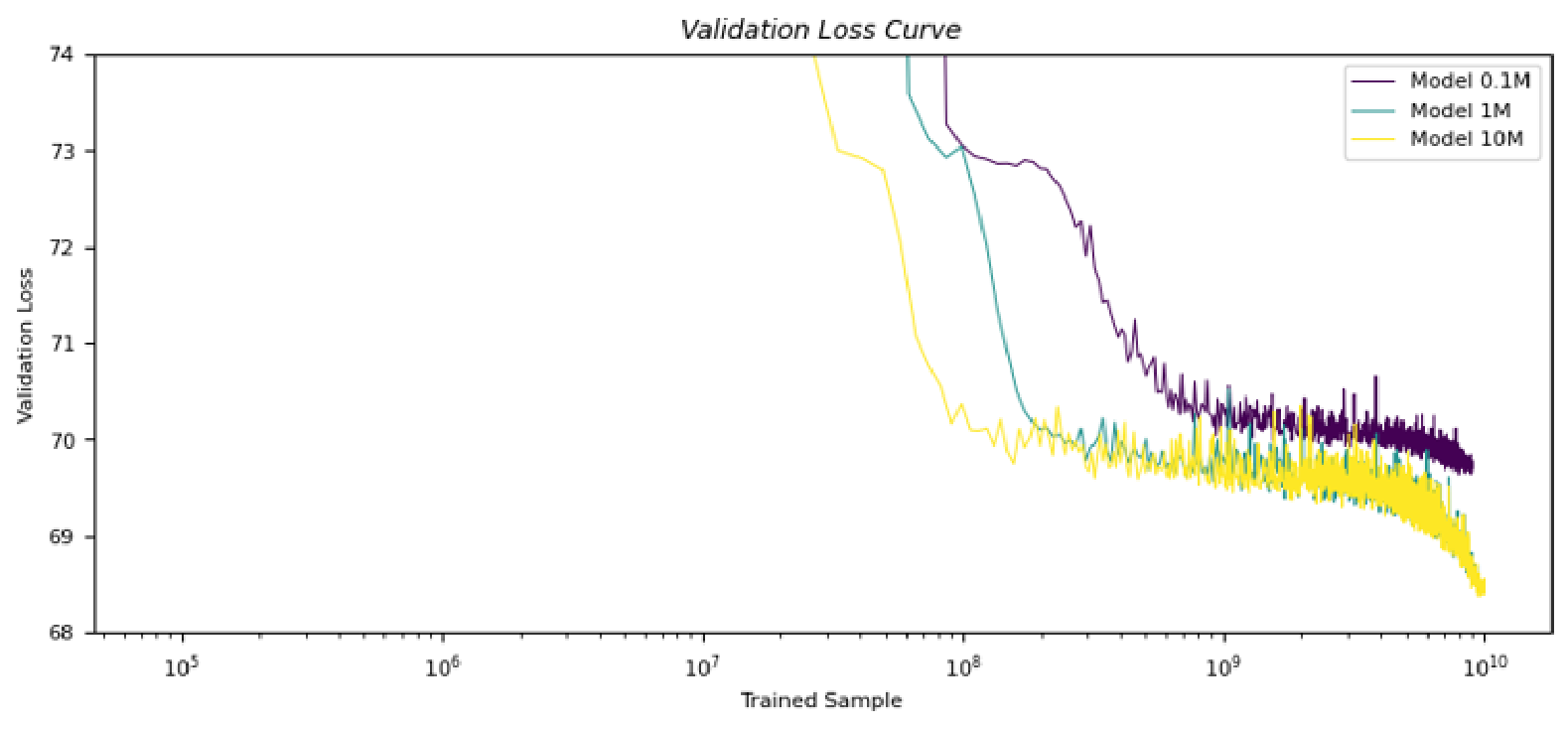}
  \caption{Variation of loss curve with model parameter count (parameter counts: 0.1M, 1M, 10M).}
  \label{fig8}
\end{figure}

Note that this paper does not emphasize the prediction accuracy of chaotic time series themselves, but rather aims to improve prediction accuracy for real-world time series. Therefore, we omit comparing chaotic time series prediction accuracy with various benchmarks.

\subsection{Scaling Law}
Based on Figure 6b, we plotted the number of training samples required to achieve consistent prediction performance - here defined as achieving a correlation coefficient of 0.1 as a benchmark for financial time series return prediction - at each fixed predictive horizon (Figure 9). This may represent a scaling law that we can achieve predictive performance at a certain level with extended predictive horizons for chaotic time series by increasing the number of training samples exponentially. For the 1000-unit time predictive horizon, while the correlation coefficient did not reach 0.1 even at the end of the 10 billion training sample period, no convergence was observed, suggesting that achieving this target is likely possible with approximately 100-200 billion training samples, consistent with the trends observed in the loss curves and correlation coefficient evolution. This scaling law bears similarity to the power-law relationship [14] where loss decreases proportionally with training computational resources and dataset size in conventional LLMs. One key difference is that no such scaling law was observed with respect to model parameters.

\begin{figure}[H]
  \centering
  \includegraphics[width=0.8\textwidth]{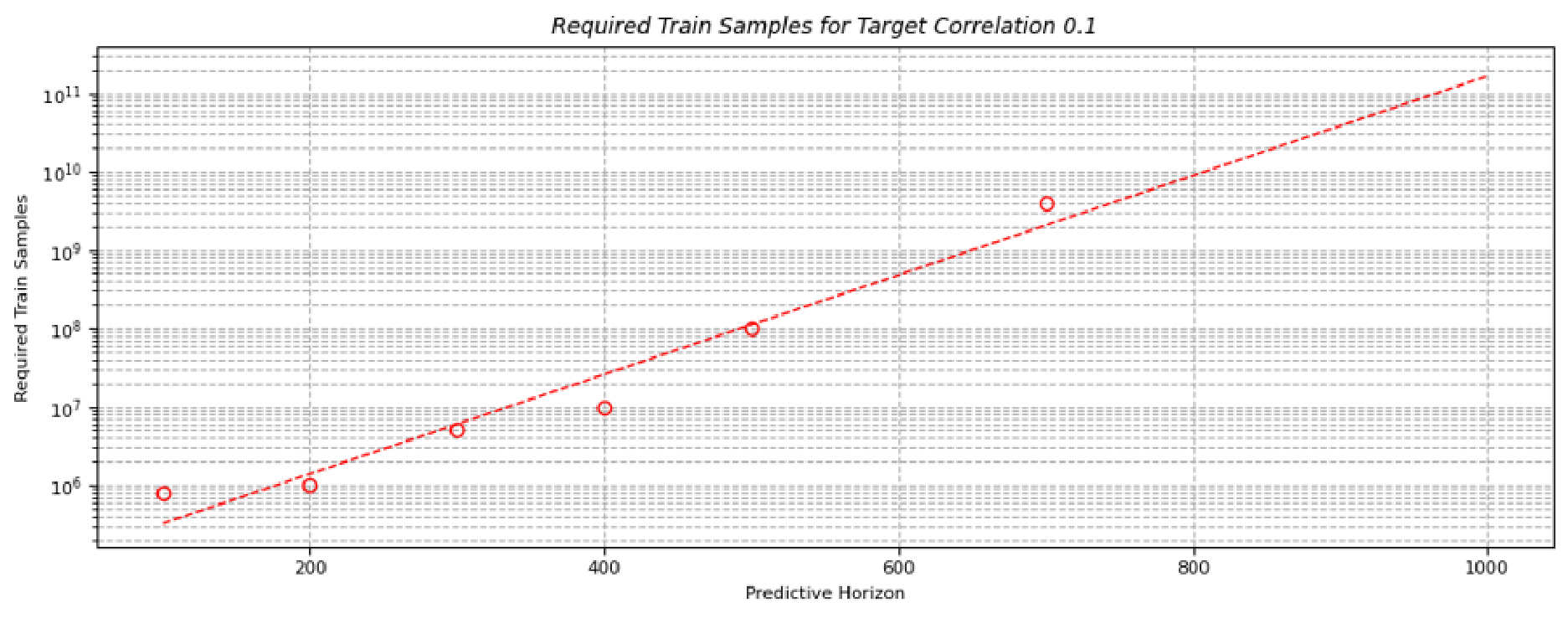}
  \caption{Scaling law for predictive horizon length in chaotic time series: The number of training samples required to achieve correlation coefficient 0.1 in time series prediction performance for the Lorenz model at each predictive horizon.}
  \label{fig9}
\end{figure}

\subsection{Prediction of Real Financial Time Series}
Figure 10 shows the balance curve resulting from trading based on the zero-shot prediction of one-step-ahead price change on test data. While trading strategies typically use the signs of predictions, in this case we adopted a strategy where predictions occurring as pulse-like outputs in high-probability situations were deemed meaningless near zero, implementing a strategy that longs the top 20th percentile (top 5\%) and shorts the bottom 5\%. Trading profits accumulated based on this strategy were evaluated as performance metrics for financial time series return prediction. Note that the calculated profits in this paper are not intended for actual trading but are used solely for assessing prediction performance. It should be noted that including transaction and execution costs would make profitability significantly more difficult to achieve at such short timeframes of less than one minute.\\
\\
At timeframes of less than one minute, price change series often exhibit strong autocorrelation, leading to variations in prediction difficulty across different timeframes. Therefore, we also verified the prediction performance of the autocorrelation model that uses the immediately preceding price change rate (the final sequence in the context) as a prediction indicator using the same method. Finally, we calculated the excess returns obtained using the pre-trained chaotic time series model at each predictive horizon relative to the absolute returns from the autocorrelation model, observing trends in prediction performance across different predictive horizons and actual time scales.\\
\\
Figure 11 shows the excess returns of models at each predictive horizon for various timeframes in financial time series. Timeframe of 15 seconds show particularly accurate prediction performance. For financial time series with extremely short timeframes like 5 seconds, the autocorrelation is very strong, causing the excess returns to be underestimated. Overall, we achieved superior results compared to statistical models like the autocorrelation model at timeframes of 15 seconds or less. Conversely, prediction performance drops sharply beyond 20 seconds. Therefore, we conclude that the 1000-unit time predictive horizon for the Lorenz model roughly corresponds to the 10-15 second timeframe of Bitcoin examined in this study, with performance degradation observed beyond this predictive horizon. Future work will investigate behavior when extending predictive horizons. Overall, we found that a 1000-unit time predictive horizon outperformed other horizons, and performance did not simply improve at predictive horizons matching the actual timeframe of real-world financial series, but rather models trained with longer predictive horizons consistently showed higher performance.

\begin{figure}[H]
  \centering
  \begin{subfigure}{0.8\linewidth}
    \centering
    \includegraphics[width=\linewidth]{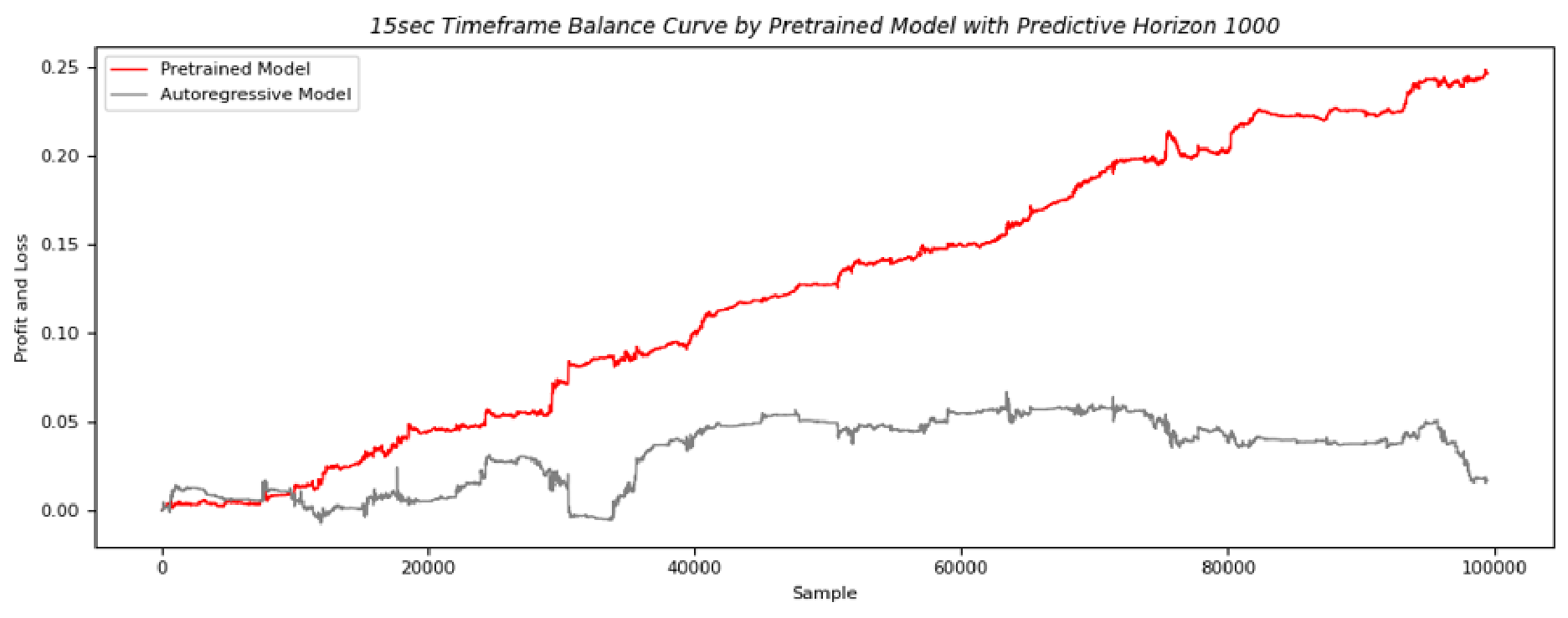}
    \subcaption*{(a)}
    \label{fig10-a}
  \end{subfigure}
  \vspace{1em}
  \begin{subfigure}{0.8\linewidth}
    \centering
    \includegraphics[width=\linewidth]{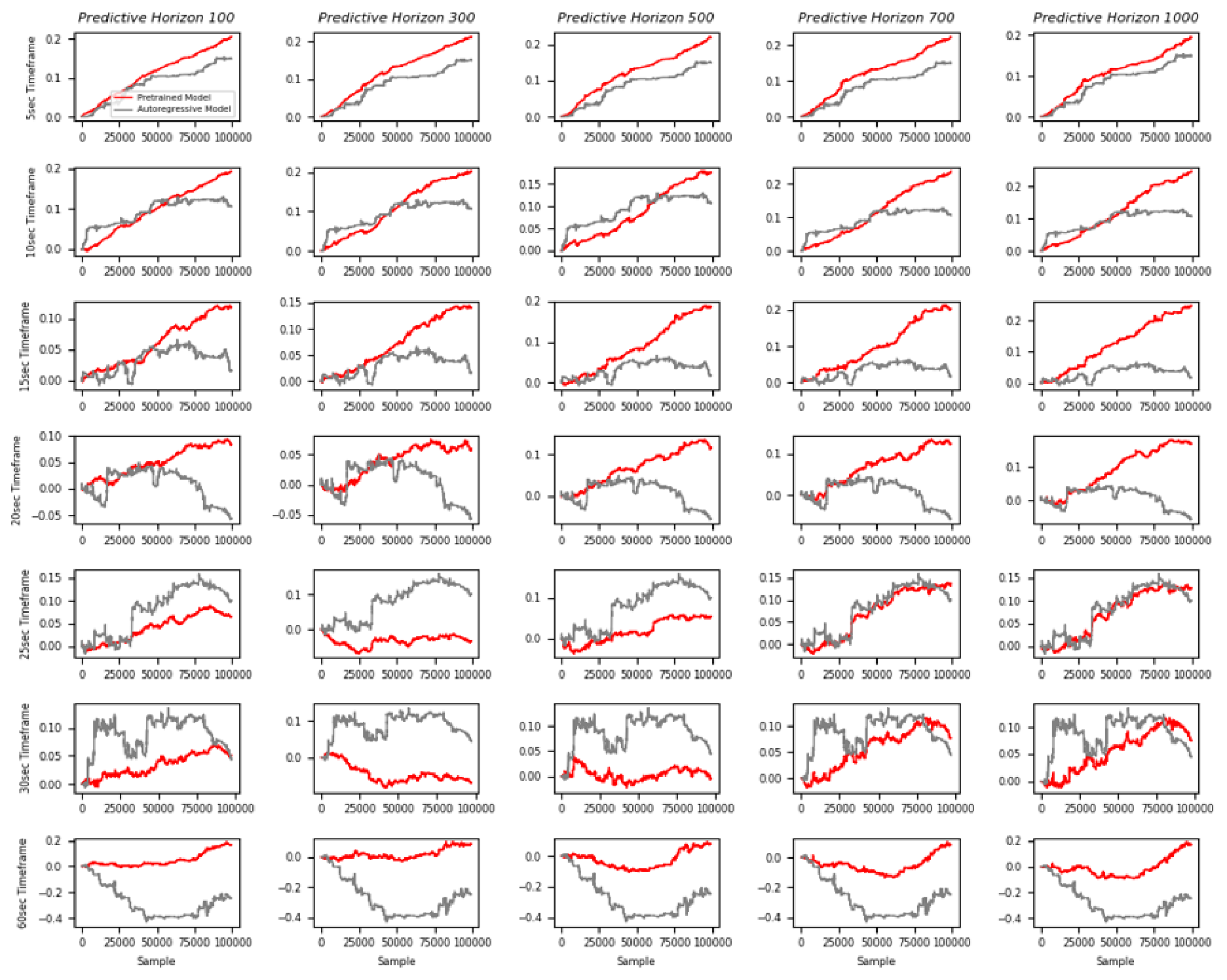}
    \subcaption*{(b)}
    \label{fig10-b}
  \end{subfigure}
  \caption{a) Representative balance curve (financial time series with timeframe of 15 seconds, predictive horizon of 1,000 unit time for the pre-trained model). b) Complete validation results. Top to bottom: financial time series with timeframe of 5 seconds, 10 seconds, 15 seconds, 20 seconds, 25 seconds, and 30 seconds; left to right: pre-trained model predictive horizons of 100, 300, 500, 700, and 1,000 unit time.}
  \label{fig10}
\end{figure}

\begin{figure}[H]
  \centering
  \includegraphics[width=0.8\textwidth]{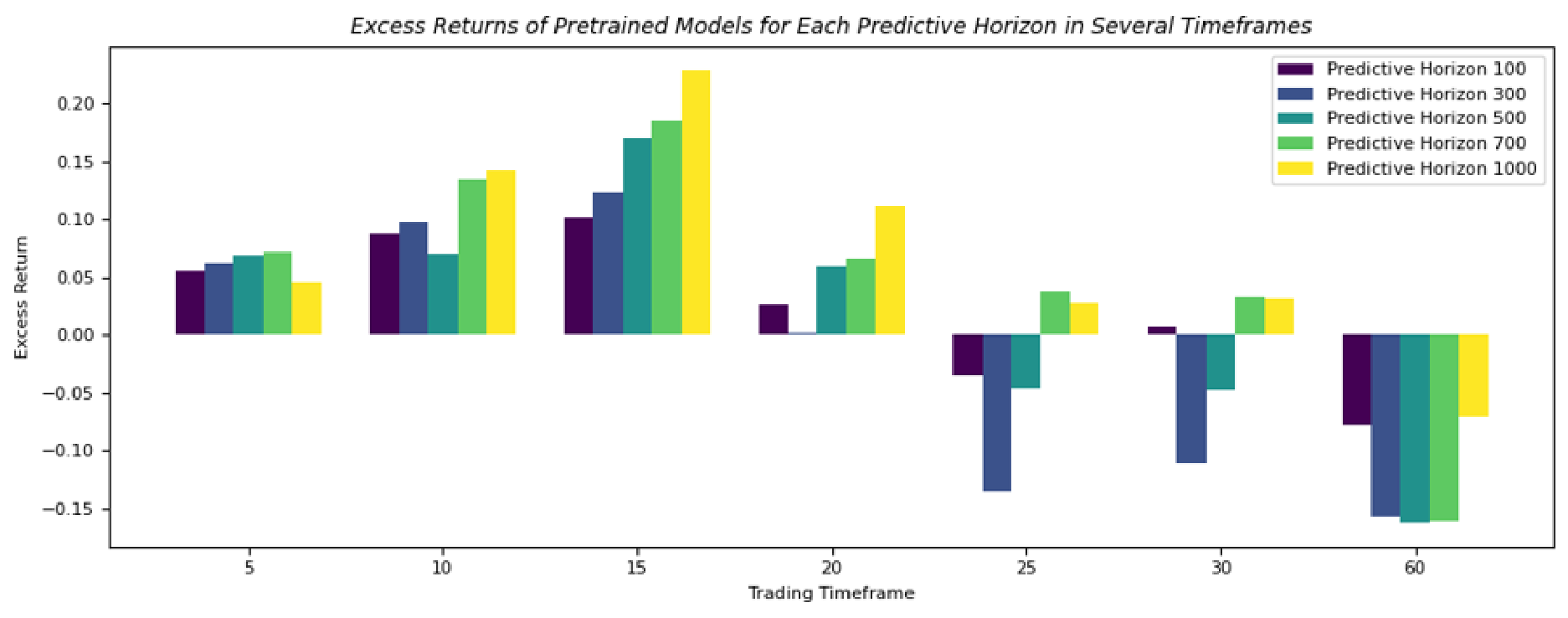}
  \caption{Model excess returns for each timeframes of financial time series across different predictive horizons.}
  \label{fig11}
\end{figure}

\begin{table}[H]
\centering
\caption{Model and autocorrelation returns along with their spreads (excess returns) for each predictive horizon across various financial time series timeframes. Red indicates the best performance, blue indicates the second-best.}
\label{tab:timeframe_horizon}
\setlength{\tabcolsep}{6pt}
\resizebox{\textwidth}{!}{
\begin{tabular}{l*{5}{cc}c}
\toprule
\multicolumn{1}{c}{\textbf{Timeframe}} &
\multicolumn{2}{c}{\textbf{Predictive Horizon 100}} &
\multicolumn{2}{c}{\textbf{Predictive Horizon 300}} &
\multicolumn{2}{c}{\textbf{Predictive Horizon 500}} &
\multicolumn{2}{c}{\textbf{Predictive Horizon 700}} &
\multicolumn{2}{c}{\textbf{Predictive Horizon 1000}} &
\multicolumn{1}{c}{\textbf{Autoregressive}} \\
\cmidrule(lr){2-3}\cmidrule(lr){4-5}\cmidrule(lr){6-7}\cmidrule(lr){8-9}\cmidrule(lr){10-11}
& \textit{Return} & \textit{Excess Return}
& \textit{Return} & \textit{Excess Return}
& \textit{Return} & \textit{Excess Return}
& \textit{Return} & \textit{Excess Return}
& \textit{Return} & \textit{Excess Return}
& \textit{Return} \\
\midrule
 5  & 0.205 & \textbf{0.056}
    & 0.211 & \textbf{0.062}
    & 0.218 & \textbf{\textcolor{blue}{0.069}}
    & 0.220 & \textbf{\textcolor{red}{0.071}}
    & 0.194 & \textbf{0.045}
    & 0.149 \\
10  & 0.193 & \textbf{0.088}
    & 0.203 & \textbf{0.097}
    & 0.176 & \textbf{0.071}
    & 0.240 & \textbf{\textcolor{red}{0.134}}
    & 0.248 & \textbf{\textcolor{red}{0.142}}
    & 0.105 \\
15  & 0.118 & \textbf{0.102}
    & 0.141 & \textbf{0.124}
    & 0.187 & \textbf{0.171}
    & 0.202 & \textbf{\textcolor{blue}{0.186}}
    & 0.246 & \textbf{\textcolor{red}{0.230}}
    & 0.017 \\
20  & 0.083 & \textbf{0.027}
    & 0.059 & \textbf{0.002}
    & 0.116 & \textbf{0.060}
    & 0.123 & \textbf{0.067}
    & 0.169 & \textbf{\textcolor{red}{0.112}}
    & 0.057 \\
25  & 0.066 & \textbf{-0.035}
    & -0.035 & \textbf{-0.136}
    & 0.054 & \textbf{-0.046}
    & 0.138 & \textbf{\textcolor{red}{0.037}}
    & 0.129 & \textbf{\textcolor{blue}{0.028}}
    & 0.101 \\
30  & 0.051 & \textbf{0.007}
    & -0.067 & \textbf{-0.110}
    & -0.004 & \textbf{-0.048}
    & 0.077 & \textbf{\textcolor{red}{0.034}}
    & 0.075 & \textbf{\textcolor{blue}{0.032}}
    & 0.043 \\
60  & 0.167 & \textbf{-0.078}
    & 0.087 & \textbf{-0.158}
    & 0.082 & \textbf{-0.163}
    & 0.084 & \textbf{-0.161}
    & 0.175 & \textbf{\textcolor{red}{-0.070}}
    & 0.245 \\
\bottomrule
\end{tabular}
}
\end{table}

\section{Conclusion}
In this paper, we hypothesized that financial time series exhibit chaotic properties at the time scales of market microstructure and even individual trader decision-making levels. We developed a method to generate data simulating financial time series through resampling of generated chaotic time series, using these as training samples. We adopted the Lorenz model as the equation for the chaotic time series, generating training samples by applying perturbations to each parameter. By expanding the resampling interval to extend predictive horizons, we performed large-scale pre-training with 10 billion training samples at each horizon, confirming a scaling law-like phenomenon that we can achieve predictive performance at a certain level with extended predictive horizons for chaotic time series by increasing the number of training samples exponentially. During the training of long predictive horizon, we observed gradual formation of the state space (attractor) as training progressed with increased training samples.\\
\\
We then created test datasets by aggregating order flow, price changes, and volume at 5, 10, 15, 20, 25, 30, and 60 second timeframes using Bitcoin trade data, and performed zero-shot predictions using pre-trained models at each predictive horizon. Verification of profitability from this simple trading strategy confirmed significant performance improvements over the autocorrelation model.\\
\\
This paper confirmed a scaling law-like phenomenon that we can achieve predictive performance at a certain level with extended predictive horizons for chaotic time series by increasing the number of training samples exponentially. If this scaling law proves robust and holds true across various chaotic models, it suggests the potential to predict near-future events by investing substantial computational resources. Reference [4] confirmed a scaling law with respect to diversity of chaotic time series used in training rather than the number of training samples, making the validity of this law across multiple chaotic models another key verification point. The author is currently conducting further training with 100 billion samples to extend predictive horizons and will verify both the validity of the scaling law and its performance on real-world time series.

\end{document}